\documentclass[final]{l4dc2020}

\title{Robust Regression for Safe Exploration in Control}
\usepackage{times}
\usepackage{wrapfig}
\usepackage{graphicx}
\usepackage{algorithm}
\usepackage{algorithmic}
\DeclareMathOperator*{\argmin}{arg\,min}

\author{%
 \Name{Anqi Liu}  \Email{anqiliu@caltech.edu} \\
 \Name{Guanya Shi} \Email{gshi@caltech.edu} \\
 \Name{Soon-Jo Chung} \Email{sjchung@caltech.edu} \\
  \Name{Anima Anandkumar} \Email{anima@caltech.edu}\\
 \Name{Yisong Yue} \Email{yyue@caltech.edu}\\
 \addr{California Institute of Technology}
 }

\begin{document}

\maketitle

\begin{abstract}%
We study the problem of safe learning and exploration in sequential control problems. The goal is to safely collect data samples from operating in an environment, in order to learn to achieve a challenging control goal (e.g., an agile maneuver close to a boundary).  A central challenge in this setting is how to quantify uncertainty in order to choose provably-safe actions that allow us to collect informative data  and  reduce uncertainty, thereby achieving both improved controller safety and optimality. To address this challenge, we present a deep robust regression model that is trained to directly predict the uncertainty bounds for safe exploration. We derive generalization bounds for learning, and connect them with safety and stability bounds in control. We demonstrate empirically that our robust regression approach can outperform conventional Gaussian process (GP) based safe exploration in settings where it is difficult to specify a good GP prior.
\end{abstract}


\section{Introduction}

% Paragraph1: Active data collection in control is important but challenging.
% Paragraph2: The key problem we need to solve is exploration under constraints
% Paragraph3: Using landing example to showcase what we are trying to do.
% Paragraph4: The proposed method and contribution

\begin{wrapfigure}{r}{0.46\textwidth}
  \centering
  \includegraphics[width=0.45\textwidth]{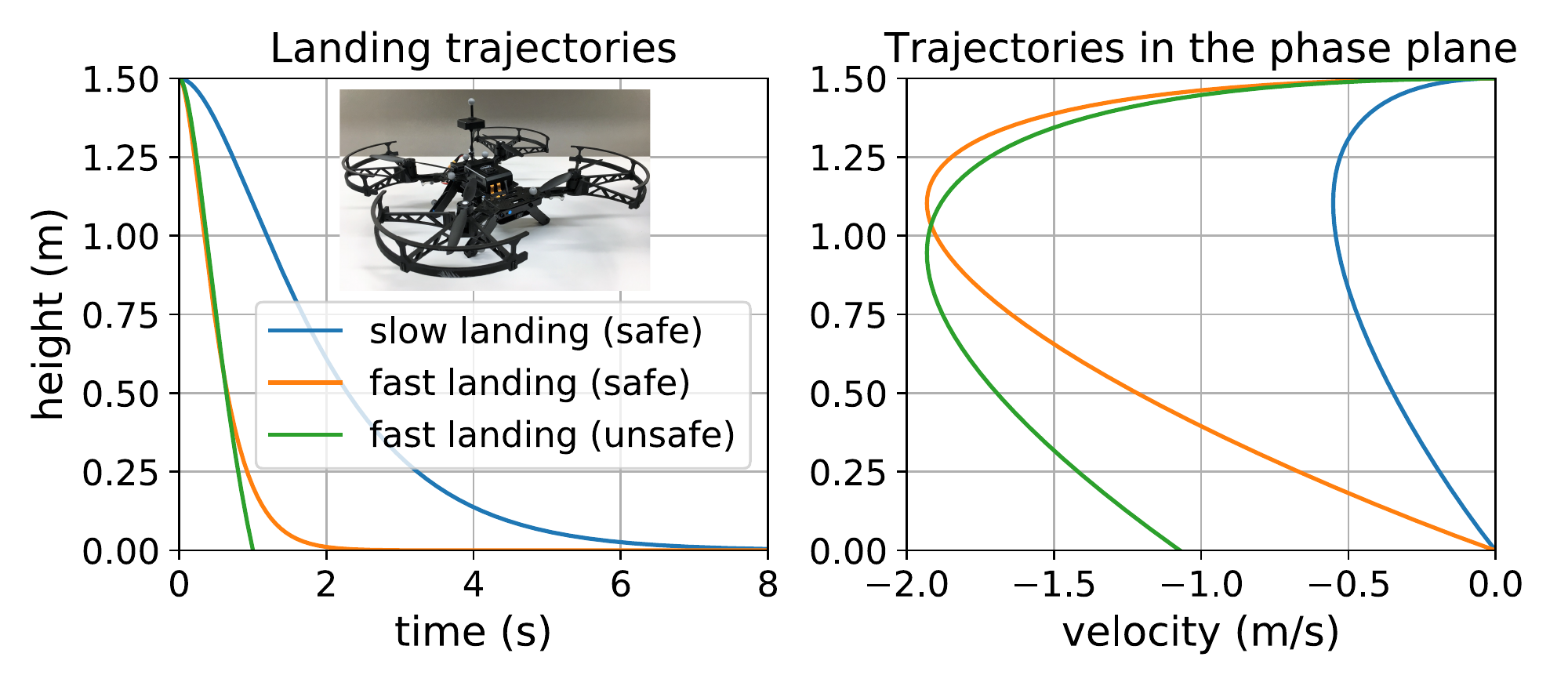}
  \label{fig:motivation}
  \vspace{-20pt}\caption{Fast drone landing}
\end{wrapfigure}

A key challenge in  data-driven design for robotic controllers is automatically and safely collecting training data.
%It is a very challenging task to learn and control a dynamical model while adaptively collecting training samples with stable controllers.
Consider safely landing a drone at fast landing speeds (e.g., beyond a human expert's piloting abilities). The dynamics are both highly non-linear and poorly modeled as the drone approaches the ground \citep{cheeseman1955effect}, but such dynamics can be learnable given the appropriate training data \citep{shi2018neural}. 
%To leverage modern learning approaches, such as deep learning, 
To collect such data autonomously, one must guarantee safety while operating in the environment, which is the problem of \textit{safe exploration}.
%Favorable approaches for modeling non-linear dynamics, as neural networks, lack theoretical guarantees for the learning performance. Moreover, safety requirements in control are brittle, and easy to break with imperfect learning from limited data. 
In the drone landing example, collecting informative training data requires the drone to land increasingly faster while not crashing.  Figure \ref{fig:motivation} depicts  an example, where the goal is to learn the most aggressive yet safe trajectory (orange), while not being overconfident and execute trajectories that crash (green); the initial nominal controller may only be able to execute very conservative trajectories (blue).

\setlength{\intextsep}{0pt}%
\setlength{\columnsep}{0pt}%
In order to safely collect such informative training data, we need to overcome two difficulties. First, we must quantify the learning errors in out-of-sample data. Every step of data collection creates a shift in the training data distribution. More specifically, our setting is an instance of covariate shift, where the underlying true physics stay constant, but the sampling of the state space is biased by the data collection \citep{chen2016robust}. In order to leverage modern learning approaches, such as deep learning, we must reason about the impact of covariate shift when predicting on states not well represented by the training set.  Second, we must reason about how to guarantee safety and stability when controlling using the current learned model. Our ultimate goal is to control the dynamical system with desired properties but staying safe and stable while data collection. The imperfect dynamical model's error translate to possible control error, which must be quantified and controlled. 

%This problem is studied as the safe exploration problem in the machine learning community. Currently, the most popular methods for safe exploration in dynamical systems are based on Gaussian processes (GPs)~\citep{rasmussen2003gaussian}, mainly due to their straightforward uncertainty quantification mechanism~\citep{akametalu2014reachability,berkenkamp2016safe,berkenkamp2017safe,fisac2018general,khalil}.  However, GPs are sensitive to model (i.e., the kernel) selection and thus not suitable for tasks that aim to gradually reach boundaries of safety sets in highly non-linear environment. In the drone example above, we would aim to gradually increase the speed of landing by choosing controls for executing higher speed trajectories, while simultaneously reducing uncertainty of dynamics model.  To satisfy the safety constraints, GP-based models will depend on well-specified kernels beforehand, which is not available in many control tasks.
% For instance, when equipped with a kernel that is overly optimistic in its ability to extrapolate, GP-based exploration can choose overly aggressive behaviors, which is highly undesirable for safety-critical tasks.   

\textbf{Our Contributions.}
In this paper, we propose a deep robust regression approach for safe exploration in model-based control. 
We view exploration as a data shift problem, i.e., the ``test'' data in the proposed exploratory trajectory comes from a shifted distribution compared to the training set. %In this way, we explicitly incorporate difference in data distribution into the learning model. 
Our approach explicitly learns to quantify uncertainty under such covariate shift, which we use to learn robust dynamics models to quantify uncertainty of entire trajectories for safe exploration. 

% Our approach builds upon robust linear regression under covariate shifts \cite{chen2016robust}, which we extend to training deep neural networks.
We analyze learning performance from both generalization and data perturbation perspectives. We use our robust regression analysis to derive stability bounds for control performance when learning robust dynamics models, which is used for safe exploration.  We empirically show that our approach outperforms conventional safe exploration approaches with much less tuning effort in two scenarios: (a) inverted pendulum trajectory tracking under wind disturbance; and (b) fast drone landing using an aerodynamics simulation based on real-world flight data~\citep{shi2018neural}.

\section{Problem Setup} %: Active Data Collection in Control under Constraints}
% Active Data Collection in Control under Constraints (Problem setup)
%       0.  Notations
% Paragraph1: We do model-based control with learning goal as partial dynamics
% Paragraph2: The nominal dynamics and learned dynamics
% Paragraph3: The controller design
% Paragraph4: The safety constraints
% Paragraph5: Goal is to have learning and control interact with each other

At a high level, our problem can be framed as a three-way interaction of: (i) learning the unmodeled, or residual, dynamics from collected data, (ii) determining whether the current learned dynamics model enables tracking a given trajectory within a safety set, and (iii) selecting trajectories for data collection that are both safe and informative, i.e., safe exploration.  In the drone landing example in Figure \ref{fig:motivation}, the residual dynamics is the ground effect that perturbs the nominal  multi-rotor model, the safety set is not crashing into the ground, and safe exploration pertains to selecting the most aggressive landing trajectory that is provably safe with the current learned dynamics model.

{\bf A Mixed Model for Robotic Dynamics.}
We consider a standard mixed model for continuous robotic dynamics \citep{shi2018neural}:
% {\small
% \begin{align}
$M(q)\ddot{q} + C(q,\dot{q})\dot{q} +G(q)-Bu = \underbrace{d(q,\dot{q})}_{\text{unknown}}$, 
% \label{eq:mixed-dynamics}
% \end{align}
% }
with generalized coordinates $q\in\mathbb{R}^n$ (and their first \& second time derivatives, $\dot{q}$ \& $\ddot{q}$), control input $u\in\mathbb{R}^m$, inertia matrix $M(q)\in\mathbb{S}^n_{++}$, centrifugal and Coriolis terms $C(q,\dot{q})\in\mathbb{R}^{n\times n}$, gravitational forces $G\in\mathbb{R}^n$, actuation matrix $B\in\mathbb{R}^{n\times m}$ and some unknown residual dynamics $d\in\mathbb{R}^n$. Note that the $C$ matrix is chosen to make $\dot{M}-2C$ skew-symmetric from the relationship between the Riemannian metric $M(q)$ and Christoffel symbols. Here $d$ is general, which potentially captures both parametric and nonparametric unmodeled terms. We aim to learn the unknown, or residual, dynamics $d(q,\dot{q})$ using machine learning models. The intuition behind this hybrid dynamical model is the sample efficiency of learning the residual should be much smaller than learning the whole model directly from data. 

{\bf Model Based Nonlinear Control.}
%The purpose of this work is to study how to explore, or adaptive collect data, to learn the residual dynamics, and so we employ standard nonlinear controller design.
To keep the ancillary design choices simple, we employ a standard nonlinear controller design \citep{shi2018neural}.
Define the reference trajectory as $\dot{q}_r=\dot{q}_g-\Lambda\Tilde{q}$, where $\Tilde{q}=q-q_g$, and the composite variable as $s=\dot{q}-\dot{q}_r=\dot{\Tilde{q}}+\Lambda\Tilde{q}$, where $\Lambda$ is uniformly positive definite. The control objective is to drive $s$ to $0$ or a small error ball in the presence of bounded uncertainty. Assuming we had a good estimate  $\hat{d}(q,\dot{q})$ of $d(q,\dot{q})$, then our controller is:
{\small
\begin{align}
u = B^\dag(M(q)\ddot{q}_r+C(q,\dot{q})\dot{q}_r-Ks+G(q)-\hat{d}(q,\dot{q})),
\label{eq:control-law}
\end{align}
}
where $K$ is a uniformly positive definite matrix, and $\dag$ denotes the Moore-Penrose pseudoinverse.

With the control law Eq.~\ref{eq:control-law}, we will have the following closed-loop dynamics:
\begin{equation}
\begin{bmatrix}M(q) & 0 \\ 0 & I\end{bmatrix}
\begin{bmatrix}\dot{s} \\ \dot{\Tilde{q}}\end{bmatrix} +
\begin{bmatrix}C(q,\dot{q})+K & 0 \\ -I & \Lambda\end{bmatrix}
\begin{bmatrix}s \\ \Tilde{q}\end{bmatrix} = 
\begin{bmatrix}d-\hat{d} \\ 0\end{bmatrix} = 
\begin{bmatrix}\epsilon \\ 0\end{bmatrix}.
\label{eq:closed-loop}
\end{equation}
where $\epsilon=d-\hat{d}$ is the approximation error between $d$ and $\hat{d}$.

{\bf Safety Requirements.}
% The state vector is denoted as $x(t)=[q(t),\dot{q}(t)]$, $x\in\mathbb{R}^{2n}$. 
For any time-varying desired trajectory, $x_g(t)=[q_g(t),\dot{q}_g(t)]$, we must certify 
%The main objective for our robotic system is to track some time-varying desired trajectory $x_d(t)=[q_d(t),\dot{q}_d(t)]$. 
safety during trajectory tracking: $x(t)\in\mathfrak{S}, \forall t$, with high probability, where $\mathfrak{S}$ is some safety set. It is obvious that $x_g(t)\in\mathfrak{S},\forall t$. However, because of unknown dynamics $d(q, \dot{q})$, the tracking error $\Tilde{x}(t)\triangleq x(t)-x_g(t)$ may be large such that $\exists t,x(t)\notin\mathfrak{S}$. 
%For example, in the drone landing problem, our desired safety requirement is the landing speed should not be higher than a upper limit.
In the drone landing example in Figure \ref{fig:motivation}, the safety set is that the vertical velocity at the point of landing should not exceed an upper limit (otherwise the drone is considered to have crash landed).

{\bf Safe Exploration.}
The ultimate goal is to identify a model (and accompanying controller) that can safely track trajectories with minimal cost.  We assume that the cost function over trajectories is known (e.g., landing as quickly as possible), but certifying safety is difficult.  The goal of safe exploration is then to select a trajectory to track that is both provably safe (with the current model) and leads to informative training data for improving safety certification.  
Our safe exploration procedure is thus to choose the lowest cost safe trajectory, which is a trajectory that lies at the boundary of the current safety set and closest to the overall minimal cost trajectory. 

\begin{figure}[bthp]
        \centering
        \setlength{\tabcolsep}{7pt}
        \begin{tabular}{c}
         \includegraphics[height=5.5cm, trim={1.5cm, 6cm, 2.5cm, 6cm}, clip]{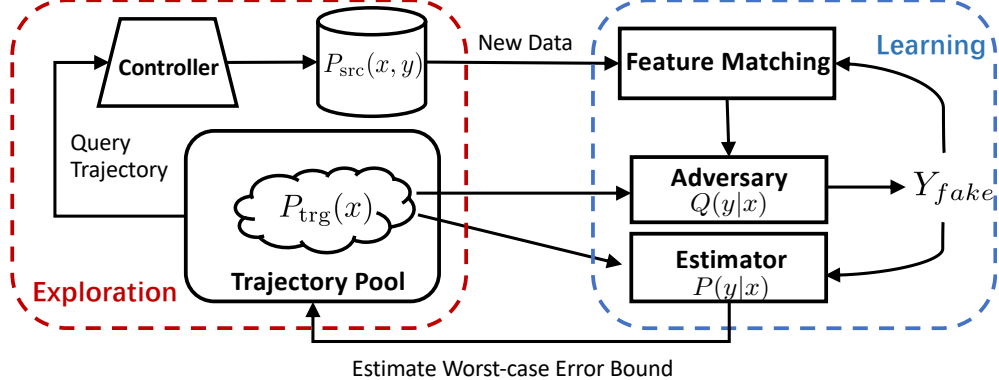}
     
        \end{tabular}
        \vspace{-0.3in}
         \caption{Our overall formulation. In the learning component, our estimator is robust to the worst-case model of the dynamics that is consistent with the observed source data, which we elaborate in Section \ref{sec:learning}. The learning and tracking error bound is then used for picking a trajectory that is safe if the worst case scenario is safe, whose details are in Section \ref{sec:safety_guarantees}.  With more source data, the worst-case model is constrained tighter along the exploration. We present the full algorithm in Section \ref{sec:exploration}.}
        \label{fig:flowchart}
    \end{figure}
     \vspace{-0.3in}

\section{Learning Residual Dynamics as Robust Regression under Covariate Shift
}
\label{sec:learning}
% Paragraph1: Learning partial dynamics is trying to map states to Fa (input variable and output variable in learning)
% Paragraph2: Covariate shift is inherent in exploration, and robustness is important
% Paragraph3: Robust regression formulation

Our learning problem is to estimate the residual dynamics $d(q,\dot{q})$ in a way that admits rigorous uncertainty estimates for safety certification.
%Our learning problem is recovering the physics that maps robot states to unknown dynamics in the model. 
The key challenge is that the training data and test data are not sampled from the same distribution, which can be framed as covariate shift \citep{shimodaira2000improving}.  
%With finite training data and potential data augmentation by safe exploration, covariate shift problem is inherent. 
Covariate shift refers to distribution shift caused by the input variables $P(x)$, while keeping  $P(y\equiv d(x)|x)$ fixed. 
%This assumption is valid in many cases, especially in dynamical systems.  
In our motivating safe landing example, there is a universal ``true'' aerodynamics model, but we typically only observe training data from a limited source data distribution $P_{\text{src}}(x)$.
Certifying safety of a proposed trajectory will inevitably cover states that are not well-represented by the training, i.e., data from a target data distribution $P_{\text{trg}}(x)$. In other words, the distribution of states in a proposed trajectory is not the distribution states in the training data.  

%After obtaining a trained model on a small initial data, new data is selected or generated according to certain evaluation criterion automatically to augment the original data, creating a potential biased dataset. The bias resides in the way of deciding whether a data point is selected or generated, which is not independent with input $x$ but rather biased towards the ones satisfying the criterion. This step is then repeated for finite number of times. The criterion usually depends on properties of the trained model in that step. Whether a data point is selected or generated only relates to the input. Therefore, it is a covariate shift setting.

% \subsection{Robust regression under covariate shifts}
{\bf General intuition.} We use robust regression \citep{chen2016robust} to estimate the residual dynamics under covariate shift. Robust regression is derived from a minimax estimation framework \citep{grunwald2004game}, where the estimator $P(y|x)$ tries to minimize a loss function on target data distribution $\mathcal{L}$, and the adversary $Q(y|x)$ tries to maximize the loss under source data constraints $\Gamma$:
{\small
\begin{align}
    \min_{P(y|x)}\max_{Q(y|x)\in \Gamma} \mathcal{L}.
\end{align}
}
Using the minimax framework, we achieve robustness to the worst-case possible conditional distribution that is ``compatible" with finite training data if the estimator reaches the Nash equilibrium by minimizing a loss function defined on target data distribution. 
% We use this robust estimator to quantify the errors in the learned dynamics and certify safety (see Section \ref{sec:safety_guarantees}). 
% Figure \ref{fig:flowchart} demonstrate our overall formulation, assuming a pool of desired trajectories is available.
% 

% The intuition is as follows. Since our data is limited, there could be multiple different distributions that are ``compatible" with data. Among all the possible compatible distributions, the adversary picks the one that maximizes our loss function. So the robustness we obtain from the framework is achieving Nash equilibrium against the worst-case possible conditional distribution that matches the data.
{\bf Technical Design Choices.} Our derivation hinges on a choice of loss function $\mathcal{L}$ and constraint set for the adversary $\Gamma$, from which one can derive a formal objective, a learning algorithm, and an uncertainty bound.
% See the full version for the complete details.
% \footnote{Full version: https://arxiv.org/abs/1906.05819.}
We use a relative loss function defined as the difference in conditional log-loss between an estimator $P(y| x)$ and a baseline conditional distribution $P_0(y| x)$ on the target data distribution $P_{\text{trg}}(x)P(y|x)$: $ \text{relative loss }  \mathcal{L}:=\mathbb{E}_{P_{\text{trg}}(x)Q(y|x)}\left[
    -\log\frac{P(y| x)}{P_0(y| x)}
    \right]$. To construct the constraint set $\Gamma$, we utilize statistical properties of the source data distribution $P_{\text{src}}(x)$:
{\small
% \begin{align}
\[\Gamma:=\{Q(y|x)| |\mathbb{E}_{P_{\text{src}}(x)Q(y|x)}[\Phi(x, y)] - {\bf c}| \le \lambda, \}\]
% \end{align}
} 
where $\phi(x,y)$ correspond to the sufficient statistics of the estimation task, and ${\bf c} = \frac{1}{n}\sum^{n}_{i = n}\Phi(x_i, y_i)$ is a vector of sample mean of statistics in the source data. 
This constraint means the adversary cannot choose  a distribution whose sufficient statistics deviate too far from the collected training data.
%This means the expectation of statistics under the adversary's choice of conditional distribution should not be far from the empirical measurement in the training data. 

The consequence of the above choices is that the solution has a
%We omit technical details of solving this problem and refer to \cite{chen2016robust} for more details. We state the conclusions here. The solution of this problem has the 
parametric form: $P(y|x) \propto P_0(y|x) e^{\frac{P_{\text{src}}(x)}{P_{\text{trg}}(x)}\theta^{T}\Phi( x, y)}$.
% where $
%     \theta = \arg\max_{\theta}\mathbb{E}_{P_{\text{trg}}(x)P(y|x)}\left[
%     \log P_{\theta}(y| x)
%     \right]
This form has two useful properties.   First, it is straightforward to compute gradients on $\theta$ using only the training data.
%We can directly compute
%the gradient of the original problem, since it is only associated with source training samples, without regularization:
% \begin{align}
%$\text{Obj}(\theta) := \mathbb{E}_{P_{\text{trg}}(x)P(y|x)}\left[
%    \log\hat{P}_{\theta}(Y| X)
%    \right],
%    % + \lambda ||\theta||^2_2, 
% $\nabla_{\theta} \text{Obj}(\theta)  = 
%     \mathbb{E}_{P_{\text{src}}(x)P_{\theta}(y|x)}[\Phi( X, y)]
%     - {\bf c} 
%     % + 2\lambda\theta. 
%     $.  
    One can also train deep neural networks by treating $\Phi(x, y)$ as the last hidden layer, i.e, we learn a representation of the sufficient statistics.
    Second, this form yields a concrete uncertainty bound (see Section \ref{sec:safety_guarantees}) that can be used to certify safety. For specific choices of $P_0$ and $\Phi$, the uncertainty is Gaussian distributed, which can be useful for many stochastic control approaches that assume Gaussian uncertainty. 
    \footnote{Our method generalizes naturally to multidimensional output setting, where we predict a multidimensional Gaussian.}.
\section{From Learning Guarantees to Tracking Guarantees}
\label{sec:safety_guarantees}
% Learning guarantees
% Interpretation of learning bounds
% Trajectory tracking guarantees based on learning bounds
% Interpretation of tracking bounds
% How to use these bounds in practice 
We demonstrate that we bound the learning errors on possible target data and further bound the tracking error. We then apply the bound to certify safety.
The proofs are in the appendix. % \footnote{Proofs in full version: https://arxiv.org/abs/1906.05819.}

{\bf Learning Guarantees.} The learning performance of robust regression approach can be analyzed from two perspectives: generalization error under covaraite shift and  perturbation error based on Lipschitz continuity.  The generalization error  reflects the expected error on a target distribution given certain function class, bounded distribution discrepancy, and base distribution. The perturbation error reflects the maximum error if target data deviates from training but stays in a Lipschitz ball. These error bounds are compatible with deep neural networks  whose Rademacher complexity and Lipschitz  constant can be controlled and measured (e.g., spectral-normalized neutral networks).
% We first establish a general form of the bounds and then derive concrete versions for both linear and deep predictors. The proofs are in the appendix.
\begin{theorem}
\label{thm:generalization}
Assume $S$ is a training set with i.i.d. data ${ x_i,..., x_n}$ sampled from $P_{\text{src}}( x)$ , $\mathcal{F}$ is a regression  function class satisfying $\sup_{x\in \mathcal{X},f,f'\in \mathcal{F}} |f(x) -f'(x)|\le M$, $\hat{\mathfrak{R}}_S(\mathcal{F})$ is the Rademacher complexity on $S$, $W$ is the upper bound of true density ratio $\sup_{ x \sim P_{\text{src}}( x)}\frac{P_{\text{trg}}( x)}{P_{\text{src}}( x)} \le W$, $\theta_y >0$ is lower bounded by $B$, the weight estimation for  the prediction $\hat{r}(x) = \frac{\hat{P}_{\text{src}}( x)}{\hat{P}_{\text{trg}}( x)}$ is lower bounded: $\inf_{ x \in S} \hat{r}(x) \ge R$, base distribution variance is $\sigma_0^2$, and $\mathfrak{\lambda}$ is the upperbound of all $\lambda_i$ among the dimensions of $\phi( x)$.  When learning a $\hat{f}\in\mathcal{F}$ on $P_{trg}(x,y)$, the following generalization error bound holds with probability  at least $1-\delta$,
{\small
\begin{equation}
    \mathbb{E}_{P_{trg}(x,y)}[(y - \hat{f}( x))^2]\notag
    \le W\left[(2RB + \sigma_0^{-2})^{-1} + \mathcal{\lambda} + 4M\hat{\mathfrak{R}}_S(\mathcal{F})+ 3M^2\sqrt{\frac{\log\frac{2}{\delta}}{2n}}\right].
    % = W\left[(2RB + \sigma_0^{-1})^{-1} + \lambda + \frac{8A^2\mathcal{X}^2}{B^2} + \frac{3A^2\mathcal{X}^2}{B^2}\sqrt{\frac{\log\frac{2}{\delta}}{2n}}\right]
\end{equation}
}
If we assume that target data samples $x$'s stay in a ball $\mathbb{B}(\epsilon)$ with diameter $\epsilon$ from the source data $S$, $\mathbb{B}(\epsilon) = \{x | \sup_{x'\in S}\|\ x - x' \|\ \le \epsilon \}$, the true function $f(x)$ is Lipschitz continuous with constant $L$, and the robust regression mean estimator $\hat{f}$ is also Lipschitz continuous with constant $\hat{L}$, then,
{\small
\begin{equation}
\sup_{x\in \mathbb{B}(\epsilon), y\sim f(x)}[(y - \hat{f}(x))^2]\le ((2RB + \sigma_0^{-2})^{-1/2}+\sqrt{\mathcal{\lambda}} + \left(L + \hat{L}\right)\| \epsilon\| )^2.
\end{equation}
}
\end{theorem}
%\textcolor{red} {It is worth noting that the upper bound of the density ratio $W$ can be controlled by choosing the target distribution carefully in the safe exploration algorithm. In other words, we can design the desired trajectories in a way such that our learning bounds are meaningful.}
The density ratio $W$ can be controlled by choosing the target distribution carefully in the safe exploration algorithm (Alg. 1).  In other words, we can design the desired trajectories to be close enough to the training set so that the resulting tracking bounds are tight enough to guarantee safety.

{\bf Tracking Guarantees.} We set $\|\epsilon\|^2=\|y-\hat{f}(x)\|^2$ to correspond with the learning bounds. The target data is set to a single proposed trajectory $x_{\text{trg}}(t)$, which means $W$ 
%  $W \le \sup_{x \sim P_{\text{src}}(x) } \frac{P_{x_{\text{trg}}(t)}(x)}{P_{\text{src}}(x)} $
 can be bounded.  
 The second option is to use a perturbation bound, where $ x_{\text{trg}}(t)\in \mathbb{B}(\epsilon)$. We emphasize that $\|\epsilon\|$ is upper bounded with $ \|\epsilon\| \le \sup_{x\in x_{\text{trg}}(t)}\|\epsilon (x)\|$ when we define target data in a specific set and use robust regression for learning dynamics.
We show $\|x(t)-x_g(t)\|\triangleq\|\Tilde{x}(t)\|$ (Euclidean distance between the desired trajectory and the real trajectory) is bounded when the error of the dynamics estimation is bounded. Again, recall that $x=[q,\dot{q}]$ is our state, and $x_g(t)$ is the desired trajectory.

\begin{theorem}
\label{thm:epsilon}
Suppose $x$ is in some compact set $\mathcal{X}$, and $\epsilon_m=\sup_{x\in\mathcal{X}}\|\epsilon\|$. Then $\Tilde{x}$ will exponentially converge to the following ball:
% \begin{equation}
$\lim_{t\rightarrow\infty}\|\Tilde{x}(t)\|=\gamma\cdot\epsilon_m$,    
% \end{equation}
where
{\small
\begin{equation}
\gamma = \frac{\lambda_{\max}(M)}{\lambda_{\min}(K)\lambda_{\min}(M)}\sqrt{(\frac{1}{\lambda_{\min}(\Lambda)})^2+(1+\frac{\lambda_{\max}(\Lambda)}{\lambda_{\min}(\Lambda)})^2},
\end{equation}
}
% where $\gamma = \frac{2\lambda_{\max}(M)}{\lambda_{\min}(M)\lambda_{\min}(Q)}\cdot\sigma_{\max}(\begin{bmatrix}
% 0 & I \\ I & -\Lambda
% \end{bmatrix})$, and $Q=\begin{bmatrix}2K & -\alpha I \\ -\alpha I & 2\alpha\Lambda
% \end{bmatrix}$ with
where $\lambda_{\max}$ denotes the maximum eigenvalue and $\lambda_{\min}$ denotes the minimum eigenvalue.
% In addition, control gain $K$ and $\Lambda$ need to satisfy $4K\Lambda\succ\alpha I$ to guarantee $Q$ is positive definite.
\end{theorem}

{\bf Integration to safe exploration.}
We can integrate 
the bounds on learning error and tracking error 
into safe exploration. Specially, if we can design a compact set $\mathcal{X}$ and find the corresponding maximum error bound $\epsilon_m =\sup_{x\in\mathcal{X}}\|\epsilon\|$ on it, we can use it to decide whether a trajectory in this set is safe or not by checking whether its worst-case possible tracking trajectory is in the safety set $\mathfrak{S}$. Then we only pick the safe trajectories with the minimum cost in data collection.

\vspace{-0.1in}
\section{Safe Exploration Algorithm}
 \label{sec:exploration}
% We are able to check the safety of the controller using Theorem \ref{thm:epsilon} when tracking a desired trajectory with a learned dynamics model. Therefore,
For simplicity,
%we design a pool of desired trajectories and use the current learned model of the dynamics to find the worst-case possible tracking trajectory for each of the desired ones.  
we maintain a finite set of candidate trajectories to select from for safe exploration; future work includes integration with continuous trajectory optimization \citep{nakka2019trajectory}.
The worst-case tracking trajectories can be computed by generating a ``tube" using euclidean distance in Theorem \ref{thm:epsilon}. We then eliminate unsafe ones and choose the most ``aggressive" one in terms of our cost function for the next iteration. Instead of evaluating the actual upper bound, we use $\beta\cdot \max_x\sigma(x)$ for measuring $\epsilon_m$ as an approximation, since it is guaranteed that the error is within $\beta\cdot \max_x\sigma(x)$ with high probability as long as the prediction is a Gaussian distribution, if the true function is drawn from the same distribution. Here $\sigma(x)$ is the standard deviation of the Gaussian distribution predicted by our robust regression algorithm. Algorithm~\ref{alg:safeexporation} describes this procedure. 
 \begin{wrapfigure}[20]{R}{0.5\textwidth}
% \vspace{25pt}
 \centering
    \begin{minipage}{0.48\textwidth}
       \begin{algorithm}[H]
        \caption{Safe Exploration for Control using Robust Dynamics Estimation}
        % \vspace{5pt}
        \label{alg:safeexporation}
        \begin{small}
        \begin{algorithmic}
          \STATE{\bfseries Input}: Pool of desired trajectories $x_g^k(t),k=1,2,....,K$; cost function $J$; robust regression model of dynamics $f$; controller $U$; safety set $\mathfrak{S}$; base distribution $\mathcal{N}(\mu_0, \sigma^2_0)$; parameter $\beta$\\
            \STATE Initialize dynamics model $f_0 = \mathcal{N}(\mu_0, \sigma^2_0)$
            \STATE Initialize training set = $\emptyset$, $f = f_0$
            \STATE {\bfseries While} $e = 1, ..., \mathcal{E}$\\
            \STATE \qquad Safe trajectory list $L$ = $\emptyset$
            \STATE \qquad {\bfseries For} $k = 1, ..., K$\\
            \STATE \qquad \qquad Predict $(\mu, \sigma^2) = f(x_g^k(t))$
            \STATE \qquad \qquad $\sigma_m = \max \sigma(x_g^k(t))$; $\epsilon_m = \beta \cdot\sigma_m $
            \STATE \qquad \qquad {\bfseries If} worst-case trajectory in $\mathfrak{S} $
            \STATE \qquad \qquad \qquad Add $x_g^k(t)$ to $L$
            \STATE \qquad Track $x_g^{*}(t) = \argmin_{x_g(t) \in L} J(x_g(t))$ to\\ \qquad collect data $x'(t)$ using controller $U$
            \STATE \qquad Add data $x'(t)$ to Training set
            \STATE \qquad Train dynamics model $f'$, $f = f' $
            \STATE {\bfseries Output}: dynamics model $f$, last desired trajectory $x_{\mathcal{E}}(t)$ and actual trajectory $x'_{\mathcal{E}}(t)$
            \end{algorithmic}
            \end{small}
      \end{algorithm}
      \end{minipage}
    \end{wrapfigure}  
    
\vspace{-0.1in}
\section{Experiments}
We conduct simulation experiments on the inverted pendulum and drone landing. We use kernel density estimation to estimate the density ratios.  We demonstrate that our approach can reliably and safely converge to optimal behavior.
We also compare  with a Gaussian process (GP) version of Algorithm \ref{alg:safeexporation}. In general, we find it is difficult to tune the GP kernel parameters, especially in the multidimensional output cases.
% using deep robust regression.
%However, a range of models underperform proposed methods in landing, indicating the difficulty to find right parameters for GP in practice. 

\paragraph{Example 1 (inverted pendulum with external wind).}
Unlike the classical pendulum model, we consider unknown external wind. Dynamics can be described as
% \begin{equation}
$ml^2\ddot{q}-mlg\sin q-u = d(q,\dot{q})$,
% \end{equation}
where $d(q,\dot{q})$ is external torque generated by the unknown wind. Our control goal is to track $q_g(t)=\sin(t)$, and the safety set is $\mathfrak{S}=\{(q,\dot{q}):|q|<1.5\}$.

% Note that the final desired trajectory is very aggressive from the beginning, in other word, without knowing external wind $d(\theta,\dot{\theta})$. 
We design a desired trajectory pool using
% \begin{equation}
$\mathcal{P}(C) = \{q_g(t)=C\cdot\sin(t),0<C\leq1\}$.
% \end{equation}
The ground truth of wind comes from quadratic air drag model.
% \angie{guanya- maybe say something here}
We use the angle upper bound in trajectory as the reward function for choosing ``most aggressive" trajectories. We use base distribution $\mathcal{N}(0, 0.5)$ to start with and $\beta = 0.5$. 
  \begin{wrapfigure}[7]{r}{0.45\textwidth}
\vspace{-2pt}
  \centering
  \begin{tabular}{cc}
  \includegraphics[height=2.5cm]{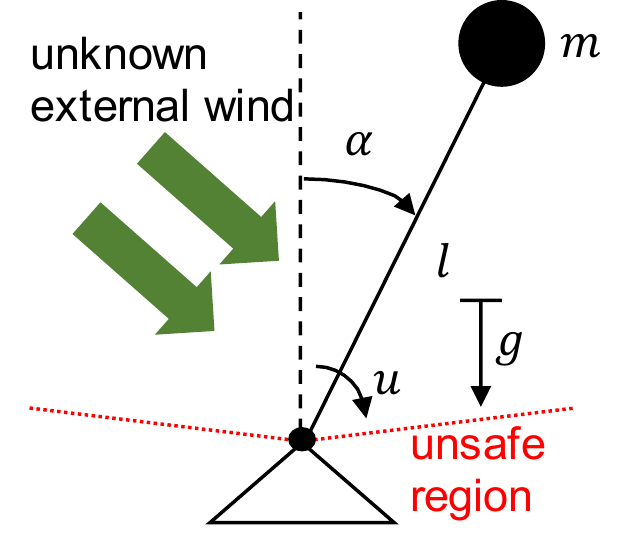}
  &\includegraphics[height=2.5cm]{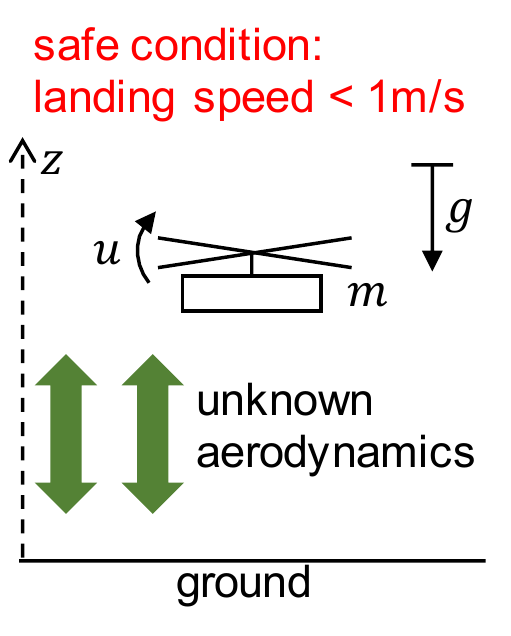}
  \end{tabular}
  \vspace{-5pt}
  \caption{Illustration of two examples}
\label{fig:example2}
\end{wrapfigure}  
\paragraph{Example 2 (drone landing with ground effect)}
We consider drone landing with unknown ground effect. Dynamics is
% \begin{equation}
$m\ddot{q} + mg - c_Tu^2 = d(q,\dot{q})$, 
% \end{equation}
where $c_T$ is the thrust coefficient.  The control goal is smooth and quick landing, i.e., quickly driving $q\rightarrow0$. The safety set is $\mathfrak{S}=\{(q,\dot{q}):\text{when}\;q=0,\dot{q}>-1\}$, i.e., the drone cannot hit the ground with high velocity. Our desired trajectory pool is
% \begin{equation}
$\mathcal{P}(C,h_g) = \{q_g(t) = e^{-Ct}(1+Ct)(1.5-h_g)+h_g,0<C,0\leq h_g<1.5\}$,
% \end{equation}
which means the drone smoothly moves from $z(0)=1.5$ to the desired height $h_d$. If $h_d=0$, the drone lands successfully. Greater $C$ means faster landing. We use landing time to determine the next ``aggressive" trajectory. The ground truth of aerodynamics comes from a dynamics simulator that is trained in~\citep{shi2018neural}, where $d(q,\dot{q})$ is a four-layer ReLU neural network trained by real flying data. We use base distribution $\mathcal{N}(0, 1)$ for robust regression and $\beta = 1$.
   \begin{figure}[bthp]
        \centering
        \setlength{\tabcolsep}{-5pt}
        \begin{tabular}{cccc}
            \includegraphics[height=3.2cm, width = 4.0cm, trim={0.2cm, 0.1cm, 3cm, 0.5cm},clip]{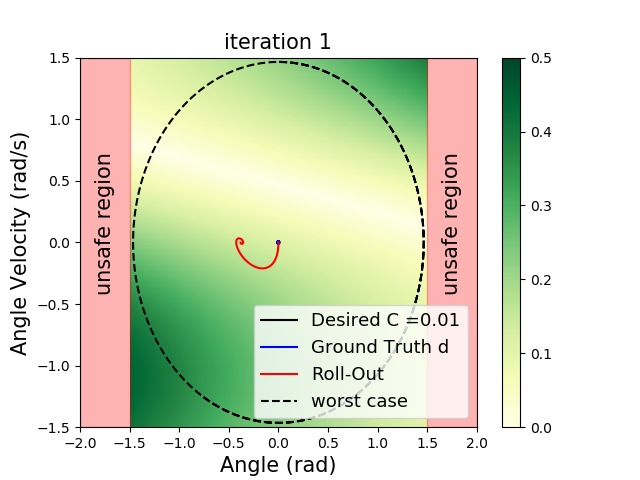}
            & \includegraphics[height=3.2cm, width = 3.9cm, trim={1.2cm, 0.1cm, 3cm, 0.5cm},clip]{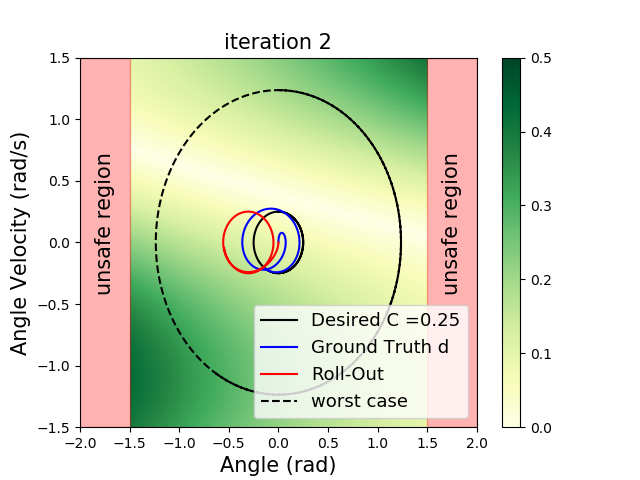} 
            & \includegraphics[height=3.2cm, width = 4.6cm, trim={1.2cm, 0.1cm, 0.5cm, 0.5cm},clip]{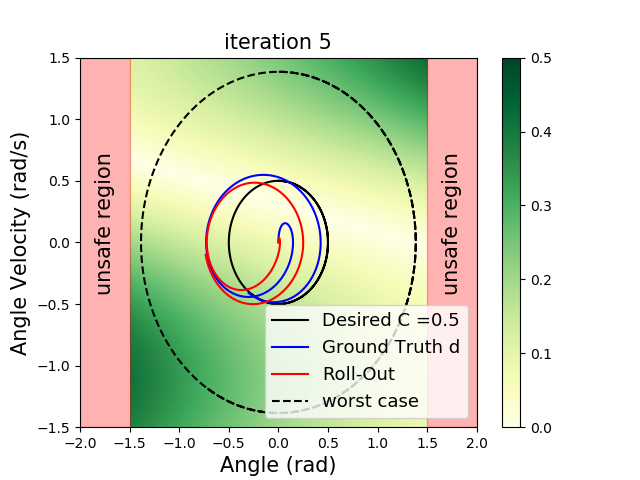}
             & \includegraphics[height=3.2cm, width = 4cm, trim={0.2cm, 0.1cm, 0.3cm, 0.5cm},clip]{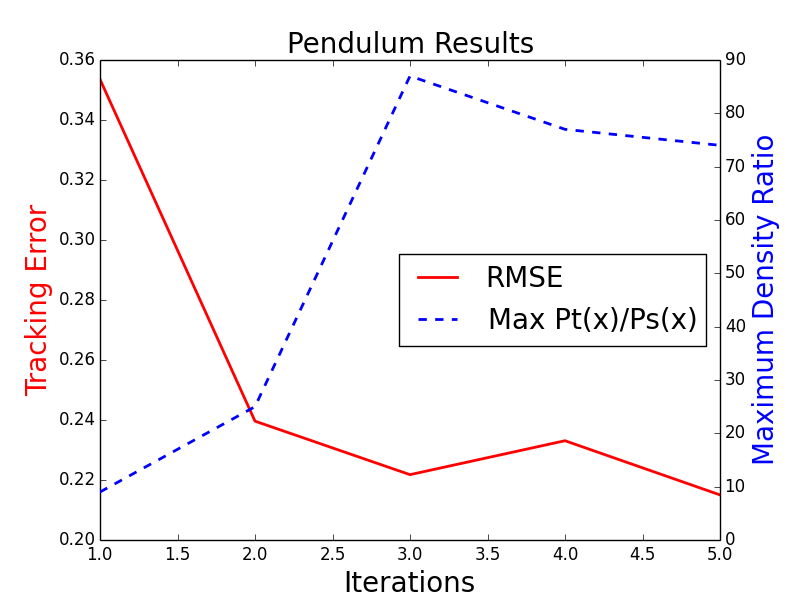}\\
            \scriptsize{(a)} & \scriptsize{(b)} & \scriptsize{(c)} & \scriptsize{(d)} \\
            \includegraphics[height=3.2cm, width = 4.0cm, trim={0.5cm, 0.1cm, 3cm, 0.8cm},clip]{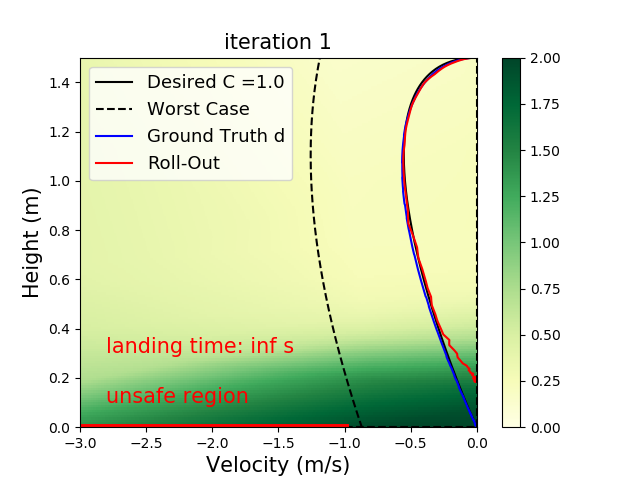}
            & \includegraphics[height=3.2cm, width = 3.9cm,trim={1.2cm, 0.1cm, 3cm, 0.8cm},clip]{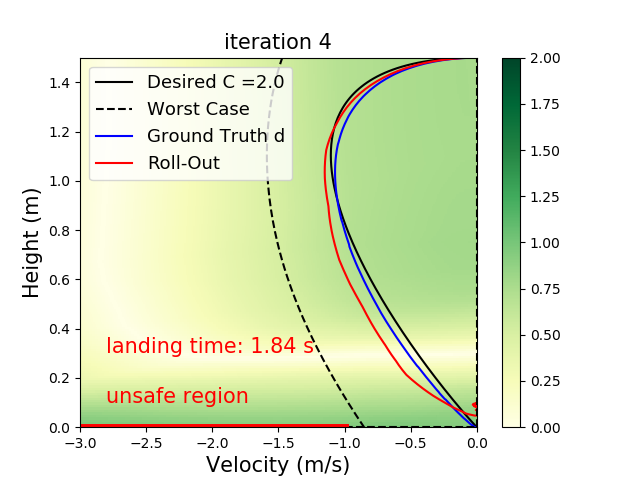} 
            & \includegraphics[height=3.2cm, width = 4.6cm,trim={1.2cm, 0.1cm, 0.5cm, 0.8cm},clip]{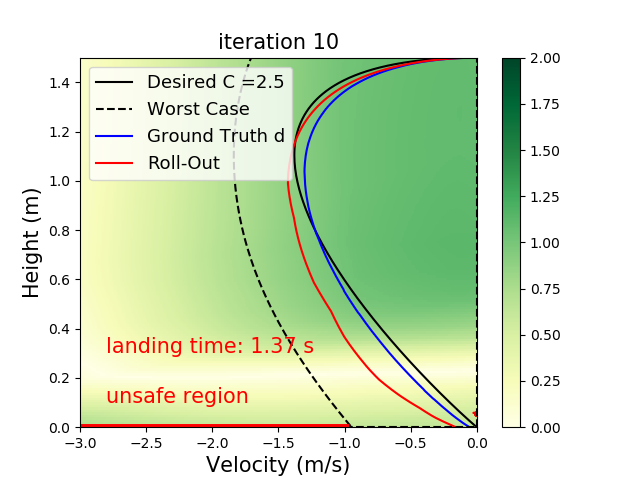}
            &\includegraphics[height=3.2cm, width = 4cm, trim={0.2cm, 0.1cm, 0.3cm, 0.5cm},clip]{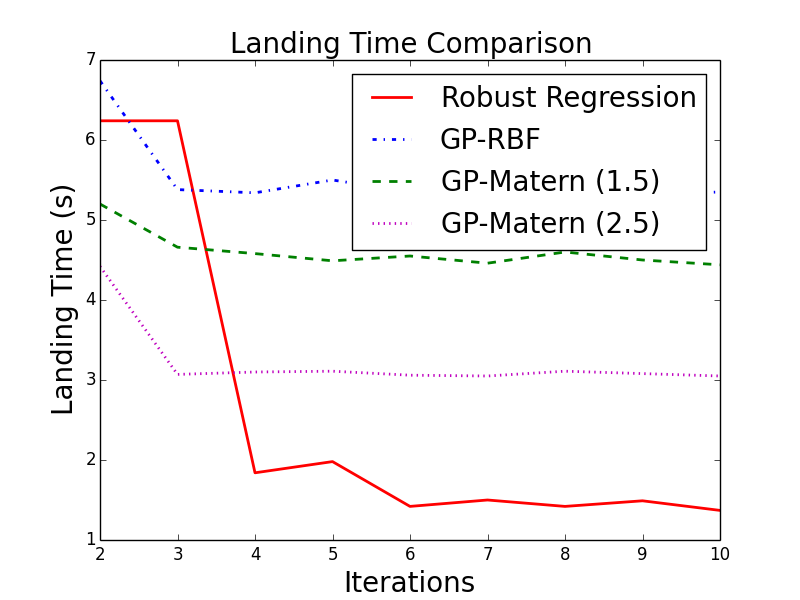}\\
            % &\includegraphics[height=3.3cm, trim={0.1cm, 0, 0, 0}, clip]{figs/Angle_pendulum.png} 
            %  &\includegraphics[height=3.3cm, trim={0.1cm, 0, 0, 0}, clip]{figs/error_bound_pendulum.png}\\
   
            \scriptsize{(e)} & \scriptsize{(f)} & \scriptsize{(g)} & \scriptsize{(h)}\\
        \end{tabular}
        \vspace{-0.2in}
         \caption{\textbf{Top Row.} The pendulum task: (a)-(c) are the phase portraits of angle and angular velocity; Blue curve is tracking the desired trajectory with ground-truth disturbance; the worst-case possible trajectory is calculated according to Theorem \ref{thm:epsilon}; heatmap is the difference between predicted dynamics (the wind) and the ground truth; and (d) is the tracking error and the maximum density ratio. 
         \textbf{Bottom Row.} The drone landing task: (e)-(g) are the phase portraits with height and velocity; heatmap is difference between the predicted ground effect) and the ground truth; (h) is the comparison with GPs in landing time.}
        \label{fig:pendulum}
        {\vspace{-0.1in}}
    \end{figure}
\paragraph{Result Analysis}
Figure~\ref{fig:pendulum}(a) to (c) and (e) to (g) demonstrate the exploration process with selected desired trajectories, worst-case tracking trajectory under current dynamics model, tracking trajectories with the ground truth unknown dynamics model, and actual tracking trajectories. Note that for landing we learn three-dimensional ground effect where $d(q,\dot{q})$ corresponds to the $z$-component, while the trajectory design and error bound computation depend on $z$-component. In both tasks, the algorithm selects small $C$ to guarantee safety at the beginning, and gradually is able to select larger $C$ values and track it while staying safe. We also demonstrate the decaying tracking error in each iteration for the pendulum task in Figure~\ref{fig:pendulum}(d). We validate that our density ratio is always bounded along the exploration.
% Results in Figure~\ref{fig:pendulum}(e) to (g) demonstrate that, because of unknown aerodynamics $d(q,\dot{q})$, $h_d = 0$ and big $C$ may not be safe at the beginning.  Starting from  conservative desired trajectories, the proposed approach is able to track more aggressive desired trajectory with big $C$ while staying safe.  
We examine the drone landing time in Figure \ref{fig:pendulum}(h) and compare against multitask GP models \citep{bonilla2008multi} with both RBF kernel and Matern kernel.
 %We compare with GP and show their landing time comparison. 
Our approach outperforms all GP models. Modeling the ground effect is notoriously challenging \citep{shi2018neural}, and the GP suffers from model misspecification, especially in the multidimensional setting \citep{owhadi2015brittleness}. Besides, GP models are also more computationally expensive than our method in making predictions. 
In contrast, our approach can fit general non-linear function estimators such as deep neural networks adaptively to the available data efficiently, which leads to more flexible inductive bias and better fitting of the data and uncertainty quantification. 
%  Additional results for the drone landing setting as well as inverted pendulum are in Appendix~\ref{app:results}. The supplemental material also contains a video demoing the results.
% \vspace{-0.1in}

\section{Related Work}
\textbf{Safe Exploration.}
Most approaches for safe exploration use Gaussian processes (GPs) to quantify uncertainty \citep{sui2015safe,sui2018stagewise,kirschner2019adaptive,akametalu2014reachability,berkenkamp2016safe,turchetta2016safe,wachi2018safe,berkenkamp2017safe,fisac2018general,khalil}. These methods are related to bandit algorithms~\citep{bubeck2012regret} and typically employ upper confidence bounds
~\citep{auer2002using} to balance  exploration versus exploitation~\citep{srinivas2009gaussian}. 
However, GPs are sensitive to model (i.e., the kernel) selection, and thus are often not suitable for tasks that aim to gradually reach boundaries of safety sets in a highly non-linear environment. In the high-dimensional case and under finite information, GPs suffer from bad priors even more  severely \citep{owhadi2015brittleness}. 
% In the drone landing example, the fastest landing trajectory is one that is just barely safe.  
%To satisfy the safety constraints, GP-based models will depend on well-specified kernels beforehand, which is not available in many control tasks.
%GP-based approaches can be sensitive to model selection.
One could blend GP-based modeling with general function approximations (such as deep learning) \citep{berkenkamp2017safe,cheng2019end}, but the resulting optimization-based control problem can be challenging to solve.  
Other approaches either require having a safety model pre-specified upfront \citep{alshiekh2018safe}, are restricted to relatively simple models \citep{moldovan2012safe}, have no convergence guarantees during learning \citep{taylor2019episodic}, or have no  safety guarantees \citep{garcia2012safe}.  
% Our work also shares some similarity with \citep{snoek2015scalable,brookes2019conditioning}, which use deep neural networks for sampling-based optimization; however, those approaches have no guarantees and so are unsuitable for safe exploration.
%Other approaches hard-code safety filters into the learning agent , but such approaches require a perfectly specified safety model upfront.
%In sequential decision making, GP-based safe reinforcement learning~\cite{berkenkamp2017safe} focus on learning dynamics in model-based reinforcement learning and explore neural network policy under stability-based safety definition. These methods rely on GP uncertainties without considering data shift. In more general safe exploration in Markov Decision Processes~\cite{garcia2012safe,turchetta2016safe,wachi2018safe}, safe exploration has been studies from an constrained optimization perspective~\cite{moldovan2012safe}, where safety constraints restrict attention to a subset of the guaranteed safe policies. Similar in continuous control, concepts like Control Barrier Functions~\cite{ames2014control} are utilized to filter controls to be safe~\cite{cheng2019end} by solving a quadratic program. However, sampling using constrained optimization is hard to analyze and it requires the safety definition to be able to represented as an affine function.
% Conservative bandit algorithm~\ref{wu2016conservative} is recently proposed to consider constraints in the reward function

\textbf{Distribution Shift.}
The study of learning under distribution shift has seen increasing interest, owing to the widespread practical issue of distribution mismatch.  
Our work is stylistically similar to \citep{liu2015shift,chen2016robust,liu2014robust,liu2017robust}, which also frame uncertainty quantification through the lens of covariate shift, although ours is the first to extend to deep neural networks with rigorous guarantees.
Dealing with domain shift is a fundamental challenge in deep learning, as highlighted by their vulnerability to adversarial inputs
\citep{goodfellow2014explaining}, and the implied lack of robustness.  Beyond robust estimation, the typical approaches are to either regularize \citep{srivastava2014dropout,wager2013dropout,simile,bartlett2017spectrally,miyato2018spectral,shi2018neural,benjamin2019measuring,corerl} or synthesize an augmented dataset that anticipates the domain shift
\citep{prest2012learning,zheng2016improving,stewart2017label}. 
We also utilize spectral normalization \citep{bartlett2017spectrally} in conjunction with robust estimation.

\textbf{Robust and Adaptive Control.} Robust control \citep{zhou1998essentials} and adaptive control \citep{slotine1991applied} are two classical frameworks to handle uncertainties in the dynamics.
% Robust control optimizes the control performance under the worst case possible disturbance via a min-max game, e.g., $\mathcal{H}_{\infty}$ control, and adaptive control online adapt unknown parameters in dynamics via environmental feedback, e.g., $\mathcal{L}_1$ adaptive control \citep{hovakimyan2010L1}. 
% There are some recent works combining learning and robust and adaptive control. For example, 
GPs have been combined with nonlinear MPC for online adaptation and uncertainty estimation \citep{ostafew2016robust}. However, robust control suffers from large uncertainty set and it is hard to analyse convergence and quantify uncertainty in adaptive control. 
Ours is the first to explicitly consider covariate shift in learning dynamics. We pick the region to estimate uncertainty carefully and adapt the controller to track safe proposed trajectory in data collection.
\section{Conclusion}
In this paper, we propose an algorithmic framework for safe exploration in model-based control.  To quantify uncertainty, we develop a robust deep regression method for dynamics estimation. %uncertainty of deep robust regression for dynamics estimation. 
Using robust regression, we explicitly deal with data shifts during episodic learning, and in particular can quantify uncertainty over entire trajectories.
%which directly predict confidence bounds under the assumption that the next trajectory to execute is the target data. 
We prove the generalization and perturbation bounds for robust regression, and show how to integrate with control to derive safety bounds in terms of stability. These bounds explicitly translates the error in dynamics learning to the tracking error in control. From this, we design a safe exploration algorithm based on a finite pool of desired trajectories. We empirically show that our method achieves superior performance than GP-based methods in control of an inverted pendulum and drone landing examples

% We prove that the proposed safe exploration algorithm converges to the optimal dynamics estimator in its function class, as well as the optimal controller for tracking optimal desired trajectories.

% Acknowledgments---Will not appear in anonymized version
\acks{Anqi Liu is supported by PIMCO Postdoctoral Fellowship at Caltech. Prof. Anandkumar is supported by Bren endowed Chair, faculty awards from Microsoft, Google, and Adobe, DARPA PAI and LwLL grants. This work is also funded in part by Caltech’s CAST and the Raytheon Company.}

\bibliography{main}
\newpage
\appendix
\section{Appendix}
\subsection{Additional Theoretical Results}

\subsubsection{Improved Bounds for Control}
\label{app:improved}
As explained in the paper, we can further improve the learning bounds in the control context when we control the target data in a strategically way. In Theorem \ref{thm:generalization}, $W$ is the upper bound of the true density ratio of this two distribution, which potentially can be very large when target data is a very different one from the source. However, we can choose our next trajectory as the one not deviate too much from the source data in practice, so that further constraining $W$ and also $\epsilon$ in Theorem \ref{thm:generalization}. We can rewrite the theorem as:
\begin{theorem}
{\bf [Improved Generalization and perturbation bounds in general cases]} Assume $S$ is a training set $S$ with i.i.d. data ${ x_i,..., x_n}$ sampled from $P_{\text{src}}( x)$ , $\mathcal{F}$ is the function class of mean estimator $\hat{f}$ in robust regression, it satisfies $\sup_{x\in \mathcal{X},f,f'\in \mathcal{F}} |f(x) -f'(x)|\le M$, $\hat{\mathfrak{R}}_S(\mathcal{F})$ is the Rademacher complexity on $S$, $W$ is the upper bound of true density ratio $\sup_{ x \sim P_{\text{src}}( x)}\frac{P_{\text{trg}}( x)}{{P_\text{src}}( x)} \le W'$, $\theta_y$ is lower bounded by $B$, the weight estimation $\sup_{ x \in S}r(x) \le R$, base distribution variance is $\sigma_0^2$, $\mathfrak{\lambda}$ is the upperbound of all $\lambda_i$ among the dimensions of $\phi( x)$, we have the generalization error bound on $P_{trg}(x,y)$ hold with probability $1-\delta$,
\begin{align}
    &\mathbb{E}_{P_{trg}({\bf x},y)}[(y - \hat{f}( x))^2]\notag
    \le W'\left[(2RB + \sigma_0^{-2})^{-1} + \mathcal{\lambda} + 4M\hat{\mathfrak{R}}_S(\mathcal{F})+ 3M^2\sqrt{\frac{\log\frac{2}{\delta}}{2n}}\right]
    % = W\left[(2RB + \sigma_0^{-1})^{-1} + \lambda + \frac{8A^2\mathcal{X}^2}{B^2} + \frac{3A^2\mathcal{X}^2}{B^2}\sqrt{\frac{\log\frac{2}{\delta}}{2n}}\right]
\end{align}
If we assume target data samples $x$'s stay in a ball $\mathbb{B}(\epsilon)$ with diameter $\epsilon'$ from the source data $S$, $\mathbb{B}(\epsilon) = {x | \sup_{x'\in S}\| x - x' \| \le \epsilon'}$ the true function $f(x)$ is Lipschitz continuous with constant $L$ and the robust regression mean estimator $\hat{f}$ is also Lipschitz continuous with constant $\hat{L}$, 
\begin{align}
&\sup_{x\in \mathbb{B}(\epsilon), y\sim f(x)}[(y - \hat{f}(x))^2]\le ((2RB + \sigma_0^{-2})^{-1/2}+\sqrt{\mathcal{\lambda}} + \left(L + \hat{L}\right)\| \epsilon'\| )^2
\end{align}
\end{theorem}
Note that in generalization bound, we can further improve the bound if we know what is the method for estimating density ratio $r$ and further relate the overall learning performance with the density ration estimation. Here, we just use $r$ as if it is a value that is given to us beforehand. 

\subsubsection{High Probability Bounds for Gaussian Distribution }
\label{app:gaussian}
In Algorithm \ref{alg:safeexporation}, we use $\beta\sigma(x)$ as our approximation of the learning error from the robust regression instead of measuring the actual learning upper bound, which is hard to evaluate. Here we give the justification. 

If the prediction from robust regression is $\mathcal{N}(\mu(x), \sigma^2(x))$, assuming true function is drawn from the same distribution, we have $Pr\{|f(x) - \mu(x)|>\sqrt{\beta}\sigma(x)\} \le e^{-\beta/2}$. Also, for a unit normal distribution $r \sim \mathcal{N}(0, 1)$, we have $Pr\{r > c\} = e^{-c^2/2}(2\pi)^{-1/2}\int e^{-(r-c)^2/2-c(r-c)}dr \le e^{-c^2/2} Pr\{r>0\} = (1/2)e^{-c^2/2}$. Therefore, for data $S$, $|f(x) - \mu(x)|\le \beta^{-1/2}\sigma(x)$ hold with probability greater than $1 - |S|e^{-
\beta\sigma/2}$.
Therefore, we can choose $\beta$ in practice and it corresponds with different probability in bounds.
\subsection{Proof of Theoretical Results}
\subsubsection{Proof of Theorem \ref{thm:generalization}}
\begin{proof}
We first prove the generalization bound using standard Redemacher Complexity for regression problems:
\begin{align}
&\mathbb{E}_{P_{\text{trg}}(x,y)}[(y - \hat{y}(x))^2]\notag\\
&=\frac{P_{\text{trg}}(x,y)}{P_{\text{src}}(x,y)}\mathbb{E}_{P_{\text{src}}(x,y)}[(y - \hat{y}(x))^2]
\qquad\qquad(\text{Covariate Shift Assumption})\notag\\
&=\frac{P_{\text{trg}}(x)}{P_{\text{src}}(x)}\mathbb{E}_{P_{\text{src}}(x,y)}[(y - \hat{y}(x))^2]\notag\\
&\le W\left[\frac{1}{n}\sum_{i=1}^n(y_i - \hat{y}(x_i))^2 + 4M\hat{\mathfrak{R}}_S(\mathcal{F})+ 3M^2\sqrt{\frac{\log\frac{2}{\delta}}{2n}}\right]\notag\\
&= W\left[\frac{1}{n}\sum_{i=1}^n (y_i - \hat{y}(x_i))^2 + 4M\hat{\mathfrak{R}}_S(\mathcal{F})+ 3M^2\sqrt{\frac{\log\frac{2}{\delta}}{2n}}\right]\notag\\
&\le W\left[\frac{1}{n}\sum_{i=1}^n |y_i^2 - \hat{y}(x_i)^2| + 4M\hat{\mathfrak{R}}_S(\mathcal{F})+ 3M^2\sqrt{\frac{\log\frac{2}{\delta}}{2n}}\right]\notag\\
&= W\left[\frac{1}{n}\sum_{i=1}^n \sigma^2(x_i) + \mathcal{\lambda} + 4M\hat{\mathfrak{R}}_S(\mathcal{F})+ 3M^2\sqrt{\frac{\log\frac{2}{\delta}}{2n}}\right]
\qquad\qquad\qquad(\text{Gradient of training vanishes:} \notag\\& \qquad\qquad\qquad\qquad\qquad\qquad\qquad\qquad\qquad\qquad\qquad\qquad\qquad y^2 - (\mu(x)^2 + \sigma^2(x)) - \lambda = 0 )\notag\\
&\le W\left[\frac{1}{n}\sum_{i=1}^n \sigma^2(x_i) + \mathcal{\lambda}+ 4M\hat{\mathfrak{R}}_S(\mathcal{F})+ 3M^2\sqrt{\frac{\log\frac{2}{\delta}}{2n}}\right]
\end{align}
where $\sup_{x\in X;f,f'\in \mathcal{F}} |h(x) -h'(x)|\le M$, $\hat{\mathfrak{R}}_S(\mathcal{F})$ is the Rademacher complexity on the function class of mean estimate, and the variance term $\sigma^2(x)$ is the empirical variance of the robust regression model and follows the sigma function \cite{chen2016robust}. This is a data-dependent bound that relies on training samples. 

We next prove the perturbation bounds. Assuming $x$ stays in a ball $\mathbb{B}(\delta)$ with diameter $\delta$ from the source training data $S$, $\mathbb{B}(\epsilon) = \{x | \sup_{x'\in S}\| x - x' \| \le \epsilon \}$, the true function $f(x)$ is Lipschitz continuous with constant $L$ and the mean function of our learned estimator is also Lipschitz continuous with constant $\hat{L}$, then we have 

\begin{align}
&\sup_{x\in \mathbb{B}(\epsilon), y\sim f(x)}(y - \hat{y}(x))^2
\le \sup_{x\in S, y\sim f(x)}(|y - \hat{y}(x)| + (L + \hat{L})\| \epsilon \| )^2 
\le \sup_{x\in S, y\sim f(x)} (|y - \hat{y}(x)| +  (L + \hat{L})\| \epsilon \| )^2 \notag\\
& \le \sup_{x\in S, y\sim f(x)} (\sqrt{|y - \hat{y}(x)|^2} + (L + \hat{L})\| \epsilon \| )^2
\le \sup_{x\in S, y\sim f(x)} (\sqrt{|y^2 - \hat{y}^2(x)|} + (L + \hat{L})\| \epsilon \| )^2\notag\\
& = \sup_{x\in S} ( \sqrt{\frac{1}{n}\sum_{i=1}^n \sigma^2(x_i) + \lambda}+ (L + \hat{L}) \| \epsilon \| )^2\notag\\
% \qquad\qquad\qquad(\text{Gradient of training vanishes:} \notag
% \end{align}
% \begin{align}
%  y^2 - (\mu(x)^2 + \sigma^2(x)) - \lambda = 0 )\notag\\
% & = \max_{i }2(\sigma^2(x_i) + \lambda) + 2(L + \hat{L})^2||\epsilon||^2\notag
& = \sup_{x\in S} ( \sqrt{\frac{1}{n}\sum_{i=1}^n \sigma^2(x_i)} + \sqrt{\lambda}+ (L + \hat{L}) \| \epsilon \| )^2
\end{align}
The last equality is due to the satisfaction of the following:
\begin{align}
    y^2 - (\mu(x)^2 + \sigma^2(x)) - \lambda = 0 
\end{align}
when gradient of robust regression vanishes \cite{chen2016robust}.
If we have an upperbound for the parameter $\theta_y > B$ and the weight estimation $r \le R$, we have
\begin{align}
   \frac{1}{n}\sum_{i=1}^n \sigma^2(x_i) \le (2RB + \sigma_0^{-2})^{-1} 
\end{align}
Therefore, the generalization bound and perturbation bounds can be written as
\begin{align}
    \mathbb{E}_{P_{\text{trg}}(x,y)}[(y - \hat{y}(x))^2] \le W\left[ (2RB + \sigma_0^{-2})^{-1} + \mathcal{\lambda}+ 4M\hat{\mathfrak{R}}_S(\mathcal{F})+ 3M^2\sqrt{\frac{\log\frac{2}{\delta}}{2n}}\right]
\end{align}
\begin{align}
\sup_{x\in \mathbb{B}(\epsilon), y\sim f(x)}(y - \hat{y}(x))^2
% \le \max_{i }2(\sigma^2(x_i) + \lambda) + 2(L + \hat{L})^2||\epsilon||^2\notag\\
\le ((2RB + \sigma_0^{-2})^{-1/2}+ \sqrt{\lambda} + (L + \hat{L})\| \epsilon\| )^2
\end{align}
\end{proof}

\subsubsection{Proof of Theorem \ref{thm:epsilon}}
\begin{proof}

Consider the following Lyapunov function:
\begin{equation}
V = s^TMs. 
\end{equation}
Using the closed-loop Eq. \ref{eq:closed-loop} and the property $\dot{M}-2C$ skew-symmetric, we will have
\begin{equation}
\frac{d}{dt}V = -2s^TKs + 2s^T\epsilon.    
\end{equation}
Note that
\begin{equation}
\frac{d}{dt}V \leq -2\lambda_{\min}(K)\|s\|^2 + 2\|s\|\epsilon_m.
\end{equation}
Using the comparison lemma \cite{khalil}, we will have
\begin{equation}
\|s\| \leq \sqrt{\frac{\lambda_{\max}(M)}{\lambda_{\min}(M)}}e^{-\frac{K}{\lambda_{\max}(M)}t}\|s(0)\| + \frac{\lambda_{\max}(M)}{\lambda_{\min}(K)\lambda_{\min}(M)}(1-e^{-\frac{K}{\lambda_{\max}(M)}t})\cdot\epsilon_m.    
\end{equation}
Therefore $s$ will exponentially converge to 
\begin{equation}
\frac{\lambda_{\max}(M)}{\lambda_{\min}(K)\lambda_{\min}(M)}\cdot\epsilon_m    
\end{equation}
Since $s=\dot{\Tilde{q}}+\Lambda\Tilde{q}$, $\|\Tilde{q}\|$ will exponentially converge to
\begin{equation}
\frac{\lambda_{\max}(M)}{\lambda_{\min}(\Lambda)\lambda_{\min}(K)\lambda_{\min}(M)}\cdot\epsilon_m.
\end{equation}
Moreover, since
\begin{equation}
\|\dot{\Tilde{q}}\|\leq \|s\| + \lambda_{\max}(\Lambda)\|\Tilde{q}\|,  
\end{equation}
$\|\dot{\Tilde{q}}\|$ will converge to
\begin{equation}
(\frac{\lambda_{\max}(M)}{\lambda_{\min}(K)\lambda_{\min}(M)}+\frac{\lambda_{\max}(\Lambda)\lambda_{\max}(M)}{\lambda_{\min}(\Lambda)\lambda_{\min}(K)\lambda_{\min}(M)})\cdot\epsilon_m.    
\end{equation}
Recall that $\Tilde{x}=[\Tilde{q},\dot{\Tilde{q}}]$. Thus finally we have the following upper bound of the error ball:
\begin{equation}
\|\Tilde{x}\|\rightarrow\frac{\lambda_{\max}(M)}{\lambda_{\min}(K)\lambda_{\min}(M)}\sqrt{(\frac{1}{\lambda_{\min}(\Lambda)})^2+(1+\frac{\lambda_{\max}(\Lambda)}{\lambda_{\min}(\Lambda)})^2}\cdot\epsilon_m.  
\end{equation}
\end{proof}

\end{document}